\documentclass[letterpaper, 10 pt, conference]{ieeeconf}  

\IEEEoverridecommandlockouts                              

\title{\LARGE \bf
Plug-and-Play Reweighting for Resilient Collaborative Decision-Making in Connected Autonomous Driving
}

\author{Jiewen Liu$^{1}$,  Rui Liu$^{2}$, Matthew Lee$^{3}$, Ming C. Lin$^{2}$, Xiaorui Liu$^{1}$, and Peng Gao$^{1}$
\thanks{$^{1}$Jiewen Liu, Xiaorui Liu and Peng Gao are with the Department of Computer Science, North Carolina State University. Email: \{jliu222, xliu96, pgao5\}@ncsu.edu.$^{2}$Rui Liu and Ming C. Lin are with the Department of Computer Science, University of Maryland, College Park. Email: \{ruiliu, lin\}@umd.edu.$^{3}$Matthew Lee is with University of North Carolina at Chapel Hill. Email: matthewlee01234@gmail.com.}
\thanks{\scriptsize
\copyright\ 2026 IEEE. Personal use of this material is permitted.
Permission from IEEE must be obtained for all other uses,
in any current or future media, including reprinting/republishing
this material for advertising or promotional purposes, creating
new collective works, for resale or redistribution to servers or
lists, or reuse of any copyrighted component of this work in
other works.}%
}

\usepackage[utf8]{inputenc}
\usepackage{booktabs}
\usepackage{array}
\usepackage{xcolor}
\usepackage{soul}
\usepackage{colortbl}
\usepackage{multirow}
\usepackage{adjustbox}
\usepackage{soul}
\usepackage{graphicx}

\usepackage[normalem]{ulem}
\usepackage{amsmath}
\usepackage{amssymb}
\usepackage{cite}

\newcommand{\PPP}{\mathcal{P}}

\newcommand{\RR}{\mathcal{R}}

\newcommand{\PP}{\mathbf{P}}

\newcommand{\hh}{\mathbf{h}}

\newcommand{\vv}{\mathbf{v}}

\newcommand{\mm}{\mathbf{m}}

\newcommand{\ff}{\mathbf{f}}
\newcommand{\qq}{\mathbf{q}}
\newcommand{\kk}{\mathbf{k}}

\newcommand{\aaa}{\mathbf{a}}
\newcommand{\pp}{\mathbf{p}}

\newcommand{\MM}{\mathbf{M}}
\newcommand{\bb}{\mathbf{b}}

\newcommand{\WW}{\mathbf{W}}

\newcommand{\zz}{\mathbf{z}}

\usepackage{makecell}

\begin{document}

\maketitle

\thispagestyle{empty}
\pagestyle{empty}

\begin{abstract}
Collaborative decision-making is a fundamental capability in multi-robot systems, such as connected autonomous vehicles.  However,  perceptual noise and adversarial attacks in collaborators can severely affect decision reliability.
Overall, existing methods typically rely on retraining with attack-specific defenses or on restrictive perturbation assumptions to improve resilience, which limits their practicality. In this paper, we propose a novel Resilient Collaborative Decision-Making (RCDM) framework that consists of an attention-based encoder for extracting individual robot perceptual embeddings and an attention-based decoder for fusing collaborator perceptions and making decisions.
To improve resilience to corrupted observations, we design a novel plug-and-play reweighting module that down-weights the influence of corrupted inputs by analyzing the consistency of neighborhood points relative to the local structure and assigning smaller weights to points that deviate strongly from the local median. This module can be seamlessly integrated into attention-based collaborative decision-making without requiring additional training.
We evaluate our method in high-fidelity simulations, considering perceptual noise and five types of attacks across diverse accident-prone scenarios. Experimental results demonstrate that our approach consistently outperforms existing methods by up to {\bf 26\%} and achieves state-of-the-art resilient performance.

\end{abstract}

\section{INTRODUCTION}

Multi-robot systems have been widely studied for decades due to their scalability, reliability and parallelism. 
To enable efficient multi-robot collaboration, a fundamental capability is collaborative decision-making, with the goal of enabling the robots to make informed decisions by leveraging knowledge shared and integrated across teammates. 
It has a variety of applications, such as multi-robot collaborative search and rescue \cite{wang2025multi,9220149}, connected autonomous driving \cite{tao_directcp}, and collaborative manufacturing \cite{11151561}.

\begin{figure}[!t]
    \centering
    \vspace{6pt}
    \includegraphics[width=0.485\textwidth]{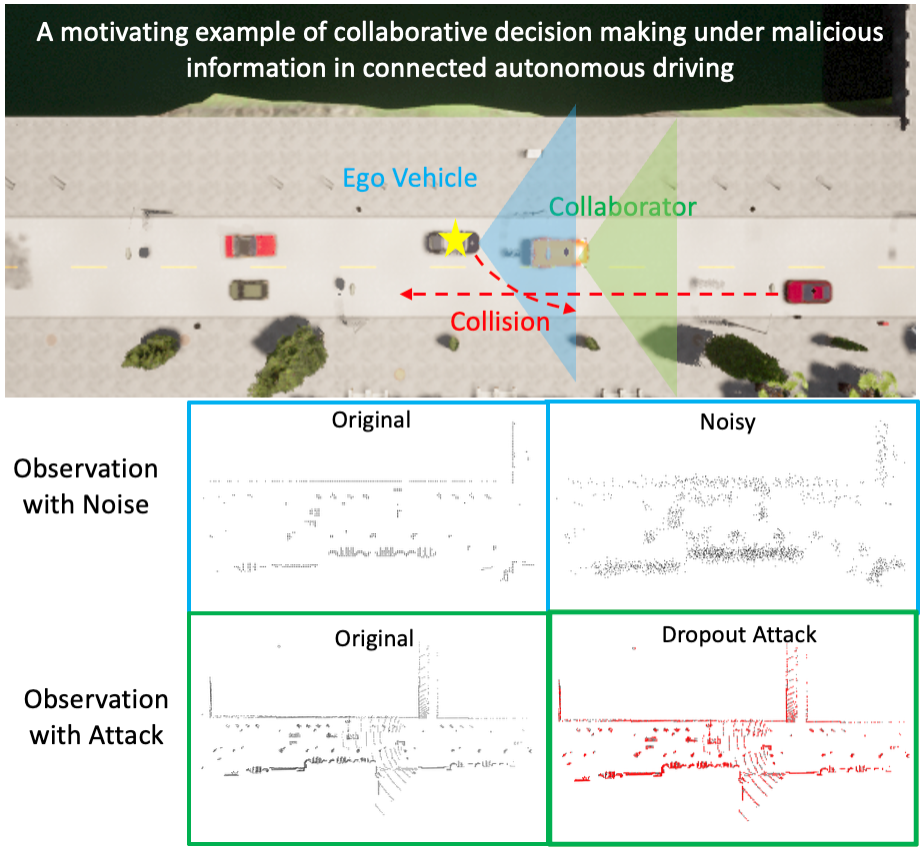}
    \caption{An illustrative example of resilient collaborative decision-making in connected autonomous driving scenarios under corrupted observations. 
    The collaborator assists the ego vehicle by sharing its perception of an oncoming red vehicle in an overtaking scenario. When the shared observations are reliable, the ego vehicle fuses the collaborator’s information with its own perception and makes a correct decision (e.g., braking) to avoid a collision.
    However, in the presence of perceptual noise or malicious attack that randomly drops points along the vertical (height) dimension, as indicated by the red dots in the Bird's Eye view (BEV), naively fusing such corrupted observations causes the ego vehicle’s perceived environment to deviate from the true state, thus leading to potential collisions.
    }
    \label{fig:motivation_figure}
    \vspace*{-1em}
\end{figure}

However, collaborative decision-making is highly vulnerable to corrupted observations, which may arise from perceptual noise in individual robots or from adversarial data transmitted by collaborators\cite{tu2021adversarial}. Such corrupted observations distort the collective understanding of the environment and can ultimately lead to unsafe decisions.
As shown in Figure~\ref{fig:motivation_figure}, the yellow collaborator assists the ego vehicle by sharing its perception of an oncoming red vehicle that may cause a serious collision. Due to perceptual noise or adversarial attacks in the collaborator’s observations, the shared information becomes inaccurate or misleading. When the ego vehicle naively merges these corrupted observations into its own perception, the resulting fused perception deviates from the true environment, causing the ego vehicle to make incorrect decisions.

Given the importance of resilient collaborative decision making under corrupted observations, many methods have been studied. First, adversarial training methods improve robustness to known perturbations by augmenting training data with attack-specific samples~\cite{wang2022art, yu2024robust}. Second, certified resilience methods provide provable guarantees under bounded perturbation assumptions, typically at the cost of additional computational overhead and restrictive attack models~\cite{liu2021pointguard,zhang2023pointcert}. Third, point removal and denoising methods attempt to filter suspicious points in raw observations based on heuristic rules, which may discard informative geometric structures~\cite{zhou2019dup,sun2021advrobust}. Finally, consensus-based methods mitigate adversarial collaborators by comparing shared observations with trusted observations and iteratively aggregating consistent information to reach consensus~\cite{RoboSAC,li2019robust}.
To sum up, {\em none of these methods can jointly improve resilience at both the individual and collaborative levels while preserving informative geometric structures and operating at inference time without retraining.}


In this paper, we propose a novel {\bf Resilient Collaborative Decision-Making (RCDM)} method. RCDM consists of an attention-based encoder that extracts embeddings from individual observations and an attention-based decoder that fuses information shared by collaborators to make final decisions, such as braking.
To improve the resilience to corrupted observations, we introduce a novel reweighting mechanism that evaluates the consistency of neighborhood points relative to the locally consistent structure and assigns smaller weights to those points deviating strongly from the local median. 
The reweighting mechanism is plug-and-play, which does not introduce any extra learning parameters and is only activated in the execution phase to enhance attention-based collaborative decision-making.


Our key contribution is the introduction of resilient collaborative decision-making under corrupted observations. Specifically,
\begin{itemize}
    \item We introduce a novel {\bf resilient collaborative decision-making} method by {\em statistically analyzing neighborhood points consistency relative to the locally consistent structure during aggregation} and {\em down-weighting the importance of corrupted observations}. Our method is resilient to both (i) perceptual noise in individual perception and (ii) attack data shared by collaborators. 

    \item We introduce a novel {\bf plug-and-play reweighting module} that can be {\em seamlessly integrated into attention-based collaborative decision-making without requiring additional training or architectural modification}. This plug-and-play design enables broad applicability across different attack models and collaborative decision-making scenarios, substantially improving deployability in real-world safety-critical systems by up to {\bf 26\%}.
\end{itemize}

\section{Related Work}

\subsection{Collaborative Decision Making in Autonomous Driving}
Existing collaborative decision making (CDM) methods can be divided into two groups, including rule-based methods and learning-based methods.
Rule-based collaborative decision-making methods rely on predefined heuristics, manually designed cost functions, or explicit coordination rules to guide collaboration among vehicles, such as decentralized lane-changing strategies based on hand-crafted decision scores\cite{nie2017decentralized}, rule-based merging with predefined temporal schedules\cite{ding2019rule}, and game-theoretic formulations that combine manually specified objectives with model predictive control for merging or roundabout scenarios \cite{hang2021cooperative,hang2021decision}.

Recently, learning-based collaborative decision-making (CDM) methods show promising performance. Graph learning approaches model traffic scenes as vehicle graphs and apply message passing to fuse neighboring states before making ego decisions \cite{chen2021graph,gao2024collaborative,klimke2022cooperative}. End-to-end collaborative driving methods directly fuse shared learned features to predict actions or trajectories, such as Coopernaut sharing BEV features among connected vehicles for decision-making in accident-prone scenarios \cite{cui2022coopernaut} and CoDriving that fuses V2X features to predict future waypoints \cite{yu2025end}. Extensions further incorporate multi-modal sensing \cite{liu2025mmcd} or learn communication policies under bandwidth constraints \cite{liu2020who2com,liu2020when2com,hu2022where2comm}. Even though these methods achieve promising performance, none of them can address corrupted observations to improve resilience of collaborative decision-making, which is particularly important in accident-prone scenarios.

\vspace{-0.3mm}
\subsection{Attacks Defense}
Attacks on LiDAR-based perception aim to manipulate point clouds or learned representations to induce erroneous predictions in downstream models. Representative attacks perturb LiDAR data through coordinate shifts, point addition, removal, or rotation, thereby disrupting local geometric structure and causing misclassification or unsafe decisions~\cite{xiang2019generating,zheng2018learning,zhao2020isometry,cao2019adversarial,li2021fooling}. Beyond single-modality manipulation, some attacks jointly corrupt LiDAR and camera inputs to exploit cross-modal dependencies in fused perception systems~\cite{hallyburton2022security,abdelfattah2021towards}. In addition, backdoor attacks embed hidden triggers during training so that specific input patterns activate malicious behavior at inference time~\cite{xiang2021backdoor,zhang2022towards,chaturvedi2024badfusion}.
Motivated by these attack models, we evaluate RCDM under three levels of increasing sophistication, including stochastic point perturbations that corrupt global geometry, targeted geometry manipulation that disrupts local structure, and gradient-based white-box attacks that exploit model internals.

Existing methods against LiDAR attacks can be generally divided into four groups.
First, adversarial-training-based methods augments training data with attack-specific samples to improve resilience~\cite{wang2022art,li2022improving, yu2024robust,ji2023benchmarking}, which cannot work well with unseen attacks.
Second, certified resilience methods provide provable guarantees under bounded perturbation assumptions, such as randomized subsampling with majority voting ~\cite{liu2021pointguard,zhang2023pointcert,commit}. However, these methods typically introduce substantial computational overhead and rely on restrictive attack models.
Third, point removal and denoising methods attempt to filter suspicious points before inference given heuristic rules~\cite{zhou2019dup,sun2021advrobust,pointcvar}, which may discard informative geometric structure.
Finally, consensus-based methods identify unreliable collaborators by comparing their information with trusted observations, such as the ego robot~\cite{RoboSAC} or neighboring collaborators~\cite{li2019robust,xue2022misspoke}, and iteratively aggregate consistent information to reach consensus.

Overall, these methods require either retraining the existing model, modeling specific attack, losing informative geometric structure, or assuming reliable observations from ego or collaborator robots. How to address the corrupted observations in both ego and collaborator robots without retraining has not been well studied yet.

\color{black}

\section{Approach}
\label{sec:approach}
\begin{figure*}[ht]
    \vspace{6pt}
    \centering
    \includegraphics[width=1\textwidth]{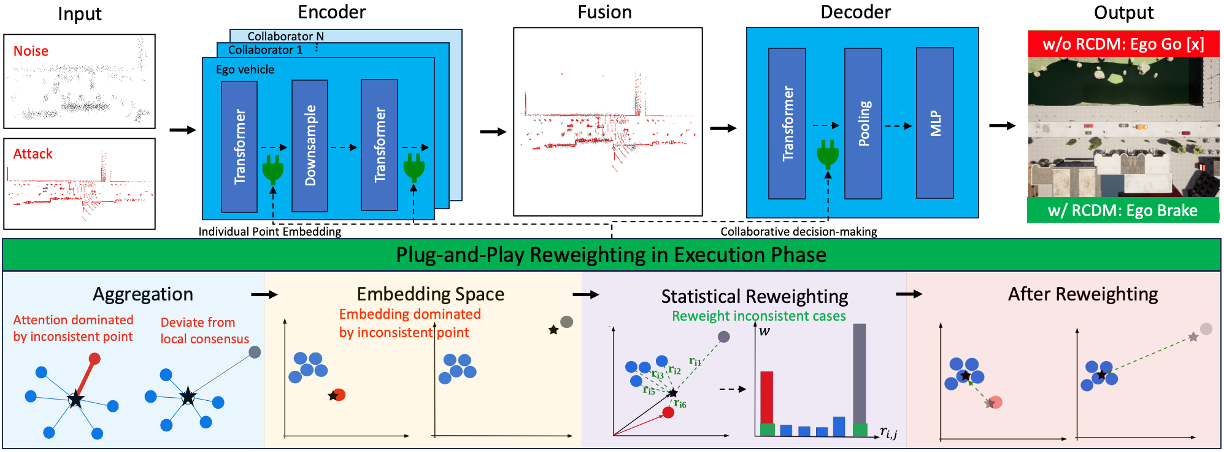}
    \caption{Overview of our RCDM method. We employ an attention-based encoder-decoder architecture for collaborative decision-making. Due to the existence of perceptual noise and malicious attacks in collaborative observations, these corrupted observations generally break the consistency of local structure during data aggregation. As a result, the generated embeddings will deviate away from the truth state. In our RCDM, by statistically analyzing the consistency of neighborhood points relative to group median, we dynamically down-weight the importance of the corrupted inputs and generate resilient embeddings for decision-making. The full reweighting process does not introduce extra learning parameters, which can be integrated with attention-based collaborative decision-making in a plug-and-play manner.}
    \label{fig:approach}
    \vspace{-3mm}
\end{figure*}

\textbf{Notation.} Matrices are represented as boldface capital letters, e.g., $\MM = \{\MM_{i,j}\}\in \RR^{n\times m}$. $\MM_{i,j}$
denotes the element in the $i$-th row and $j$-th column of $\MM$.
Vectors are denoted as boldface lowercase letters $\vv \in \RR^{n}$ and scalars are denoted as lowercase letters.

\subsection{Problem Definition}
We discuss our RCDM approach that enables resilient collaborative decision-making under corrupted observations. The overview of RCDM is illustrated in Figure \ref{fig:approach}. 
Assume that we have $n$ vehicles with one ego vehicle and $n-1$ collaborators. Each vehicle provides a LiDAR point cloud $\PPP = \{\pp_i\}$ and its velocity $v$. 
We first design an encoder $\{\hh_i\} = \psi(\PPP)$ to extract the embeddings $\{\hh_i\}$ of points in the point cloud. Then, each vehicle  down-samples the raw points, denoted as $\{\pp_i\}^m$, where $m$ denotes the number of down-sampled points. Then, the collaborator vehicles  share their down-sampled points $\{\pp_i\}^m$ and corresponding point embeddings $\{\hh_i\}^m$ with ego vehicle.

Once the ego vehicle receives these from its collaborators, it  converts the collaborators' point clouds into ego coordinates. The transformation matrix can be obtained from the GNSS or HD map. Then, the ego vehicle merges all shared points, the merged point cloud is defined as $\PPP' =\{\pp_i\}^{m^\prime}$, where $m^\prime$ denotes the number of merged points. Given the merged points $\PPP'$ and their associated embeddings $\{\hh_i\}^{m^\prime}$, we further design a decoder $y = \phi(\PPP', \{\hh_i\}^{m^\prime})$ to make the decision $y$ for the ego vehicle, where $y=1$ indicates the ego vehicle applying brake decision and $y=0$ corresponds to maintaining motion without braking. 

In real-world environments, malicious attacks or noisy perception in both the ego vehicle or its collaborators can severely undermine the reliability of collaborative decision-making, especially in accident-prone scenarios. In this paper, we mainly address two sub-problems.
\begin{itemize}
    \item \textbf{Individual point embedding under perceptual noise}. Each individual vehicle generates reliable feature embeddings from noisy perception, especially in accident-prone scenarios where accurate representations are critical for supporting timely braking decisions to avoid collisions.
    \item \textbf{Collaborative decision-making under malicious attacks}. The capability of connected vehicles to maintain resilient collaborative decision-making under malicious attacks (e.g.,  deliberate manipulation of point locations), where a single corrupted message may propagate through perception fusion, thus leading to unsafe decisions.
\end{itemize}

\subsection{Individual Point Embedding under Perceptual Noise}
In real-world scenarios, LiDAR points often suffer from noisy sensing, which significantly degrades the expressiveness of individual perceptual embeddings. To address this challenge, we design a novel attention-based encoder that analyzes the statistics of LiDAR points during data aggregation and reduces the importance of noisy data that deviate from the group median in the embedding space, thus generating resilient feature embeddings.


Specifically, we generate the embedding of each point as $\ff_i = \WW\pp_i$, where $\WW$ is a learnable matrix.
Then, we compute the query, key and value embeddings, which are defined as follows:
\begin{equation}
    \qq_i = \WW^{q}\ff_i, \quad
    \kk_j = \WW ^{k}\ff_j, \quad
    \vv_j = \WW^{v}\ff_j
    \label{eq: qkv}
\end{equation}
where $\WW^{q}, \WW ^{k}$ and $\WW ^{v}$ denote learnable matrices.  $\ff_j$ is the embedding of one of the $K$ nearest neighbors of $\pp_i$.
To encode the geometric cues into the embedding, we compute the positional embedding of the $i$-th point with respect to its nearest neighbors, which is defined as $\Delta_{i,j}=\mathrm{MLP}(\pp_i-\pp_j)$, where $\Delta_{i,j}$ encodes the relative distance between pairs of points. 
Then, we compute the attention as $\aaa_{i,j} = \mathrm{MLP}(\qq_i - \kk_j + \Delta_{i,j})$. To normalize the attention, we further apply softmax on it, defined as
\begin{equation}\label{eq:alpha}
    \alpha_{i,j} = \frac{\exp(\aaa_{i,j})}{\sum_{j' \in \mathcal{N}(i)} \exp(\aaa_{i,j'})}
\end{equation}
where $\exp$ denotes the exponential operation and $\mathcal{N}{(i)}$ denotes the neighbors of the $i$-th point. 
The final embedding of the $i$-th point is defined as $\hh_i$, which is computed by aggregating the neighbor embeddings weighted by attention:
\begin{equation}
\hh_i = \sum_{j\in\mathcal{N}(i)} {\alpha}_{i,j}\, (\vv_j + \Delta_{i,j}).
\label{eq:embed}
\end{equation}

Due to the existence of noise in $\PPP$, the computed embedding $\hh_i$  can be distorted by unreliable neighbors during aggregation, thus leading to inaccurate embedding.
Motivated by robust statistics~\cite{beaton1974fitting,bloomfield1983least, huber1973robust, zhang2010nearly, hou2024protransformer}, we design a novel plug-and-play reweighting module to suppress outliers during embedding aggregation, which can be integrated into the encoder $\psi$ during the execution phase.
Specifically,
we first compute the Euclidean distance between the value embedding $\vv_i$ and the aggregated embedding $\hh_i$, which is defined as 
\begin{equation}\label{eq:center}
r_{ij} \;=\; \big\|\vv_{j}-\hh_i\,\big\|_2 ,
\end{equation}
This distance $r_{i,j}$ indicates how far each neighbor deviates from the aggregated embedding. Based on the set of distances $\{r_{i,j}\}_{j \in \mathcal{N}(i)}$ across all neighbors of the $i-th$ node,
we compute  reference statistics as follows, 
\begin{align}
c &= \operatorname{median}_{j \in \mathcal{N}(i)} r_{i,j} \\
d &= \max\!\Big(\operatorname{median}_{j \in \mathcal{N}(i)} \lvert r_{i,j}-c\rvert,\;\varepsilon\Big).
\label{eq:center-scale}
\end{align}
where $c$ denotes the median of distances set $\{r_{i,j}\}^m$ across all neighbors of the $i$-th node, $d$ denotes the median absolute deviation around $c$, and $\epsilon$ is a hyperparameter, which serves as a lower bound to avoid division by zero in subsequent normalization. We further normalize the deviation as follows:
\begin{equation}
u_{ij} \;=\; \frac{r_{ij}- c}{kd}, \qquad k>0,
\label{eq:std}
\end{equation}
where $k$ denotes a hyperparameter that controls the sensitivity of the normalization. A smaller $k$ makes the interval around 
$c$ narrower, so even moderate deviations from $c$ lead to larger standardized scores, while a larger $k$ broadens the interval and treats such deviations as less significant. Given the normalized deviation, we compute the reweighting factor as 
\begin{equation}\label{eq:reweight}
w_{i,j} = \left( \max\{\,1-|u_{ij}|^2,\,0\} \right)^2 
\end{equation}
where $w_{i,j}$ denotes the reweighting factor. The reweighting factor $w_{i,j}$ uses the normalized deviation $u_{i,j}$, which is computed from each neighbor’s distance $r_{i,j}$ relative to the group median $c$, so that neighbors close to the group median receive weights near $1$ while those deviating significantly are downweighted toward $0$.
Intuitively, $w_{i,j}$
evaluates each neighbor’s deviation relative to the group median distance, which assigns high weights to consistent neighbors and weakens the influence of outliers that break the consistency of local structure. This leads to embeddings from noisy points receiving very small or even zero weights, effectively filtering them out during aggregation.
Importantly, the reweighting factor $w_{ij}$ can be seamlessly applied in a plug-and-play manner during execution, since it only requires an additional step of computing $w_{ij}$ and reweighting the already trained attention without modifying any network parameters. Formally, the reweighted attention is computed as follows:
\begin{equation}\label{eq:hat_alpha}
    \hat{\alpha}_{i,j} = \frac{w_{ij}\exp(\aaa_{i,j})}{\sum_{j' \in \mathcal{N}(i)} w_{ij'}\exp( \aaa_{i,j'})}
\end{equation}
By emphasizing neighbors whose distances are close to the group median and down-weighting those that deviate strongly, the attention captures the importance of points for decision-making.
The final embedding of the $i$-th point with the reweighting update is defined as follows:
\begin{equation}
\hh_i = \sum_{j\in\mathcal{N}(i)} \hat{\alpha}_{i,j}\, (\vv_j + \Delta_{i,j}).
\label{eq:vanilla-pool}
\end{equation}
The final embedding $\hh_i$ preserves the contribution of reliable neighbors and filters out the abnormal ones.
Once each vehicle generates its own embeddings, they share those points $\{\pp_i\}^m$ and associated embeddings $\{\hh_i\}^m$ among all connected vehicles, where $m$ denotes the number of down-sampled points.

\subsection{Collaborative Decision-Making under Malicious Attacks}
Once the ego vehicle receives the shared point clouds from the collaborators, it will first merge all the points with respect to its own coordinate. Then, the final decision is made through the decoder $\phi(\PPP', \{\hh_i\}^{m^\prime})$.
However, the shared information can be corrupted by perceptual noise or even intentionally manipulated by malicious attackers. Such unreliable shared information may mislead the ego vehicle’s decision-making process, thus creating significant safety issues.

To design a resilient decoder, we first compute the query, key and value embeddings of the $i$-th point in the merged point cloud $\hat{\PPP}$, according to the similar equations defined in Eq. (\ref{eq: qkv}).
Then, the attention is computed as $\bb_{i,j} = \mathrm{MLP}(\qq^\prime_i - \kk^\prime_j + \Delta^\prime_{i,j})$, where $\bb_{i,j}$ denotes the attention for decoder and $\Delta^\prime_{i,j}= \mathrm{MLP}(\pp_i-\pp_j)$ denotes the positional embedding. The normalized attention is computed as 
\begin{equation}\label{eq:beta}
    \beta_{i,j} = \frac{\exp(\bb_{i,j})}{\sum_{j' \in \mathcal{N}(i)} \exp(\bb_{i,j'})}
\end{equation}
where $\beta_{i,j}$ captures the geometric relationships of the $i$-th and $j$-th points in the merged point cloud.  However, due to the malicious collaborative information caused by attacks or perceptual noise, the attention in Eq. (\ref{eq:beta}) may assign high weights to corrupted collaborators, causing their misleading embeddings to dominate the final decision-making.
Similarly, we compute the reweighting factor $w_{i,j}$ as defined in Eqs. (\ref{eq:center} - \ref{eq:reweight}). 
Then the updated attention for collaborative decision-making is defined as
\begin{equation}\label{eq:hat_beta}
    \hat{\beta}_{i,j} = \frac{w_{ij}\exp(\bb_{i,j})}{\sum_{j' \in \mathcal{N}(i)} w_{ij'}\exp(\bb_{i,j'})}
\end{equation}
The final embedding for collaborative decision-making is computed as:
\begin{equation}\label{eq:decision}
    \zz_i = \sum_{j\in\mathcal{N}(i)} \hat{\beta}_{i,j} \left(\vv^\prime_j + \Delta^\prime_{i,j}\right)
\end{equation}
where $\zz_i$ denotes the final embedding of the $i$-th point in the merged point cloud $\hat{\PP}$.
The embedding $\zz_i$ captures geometric cues of all the points obtained by the ego and collaborator vehicles. It can weaken malicious inputs that deviate strongly away from the group median,
meanwhile preserving useful geometric cues for collaborative decision-making. 
Then we concatenate all the point embeddings and the ego vehicle's velocity embedding $\vv$, defined as $\mm = concat(\{\zz_i\}^{m^\prime}, \vv)$.
The final decision is made by $y = \mathrm{MLP}(\mm)$, where $y=1$ indicates the ego vehicle applying brake decision and $y=0$ corresponds to maintaining motion without braking.

\subsection{Unified Training and Plug-Play Execution}
To train our \textbf{RCDM} consisting of an encoder $\psi$ and a decoder $\phi$, we use the Binary Cross Entropy Loss (BCE), which is defined as 
\begin{equation}
\mathcal{L}_{\text{BCE}}
=  -(y \log \hat{p} + (1-y)\log(1-\hat{p}))
\end{equation}
where $\hat{p}$ denotes the predicted braking probability and $y \in \{0, 1\}$ denotes the ground-truth, 
with $y=1$ indicating that the ego vehicle should brake and $y=0$ 
indicating that it should maintain motion.
During the training, RCDM is optimized end-to-end on clean data without noise and attacks. Thus, the reweighting factor defined in Eqs. (\ref{eq:center} - \ref{eq:reweight}) is deactivated. Thus, the encoder attention $\hat{\alpha}$ in Eq. (\ref{eq:hat_alpha}) is downgraded to $\alpha$ in Eq. (\ref{eq:alpha}). Similarly, the decoder attention $\hat{\beta}$ in Eq. (\ref{eq:hat_beta}) is downgraded to $\beta$ in Eq. (\ref{eq:beta}).

During the execution, the reweighting factor is activated in both encoder and decoder to enhance resilience against noisy or malicious attacks. 
By activating the reweighting factor, the same trained parameters are reused, and only the additional reweighting step is applied.
This plug-and-play design allows the trained model to operate without retraining and provides resilience against malicious inputs while incurring only minor computational overhead.

\section{Experiment}
\subsection{Experimental Setup}
\begin{figure}[t]
    \vspace{6pt}
    \centering
    \includegraphics[width=\columnwidth]{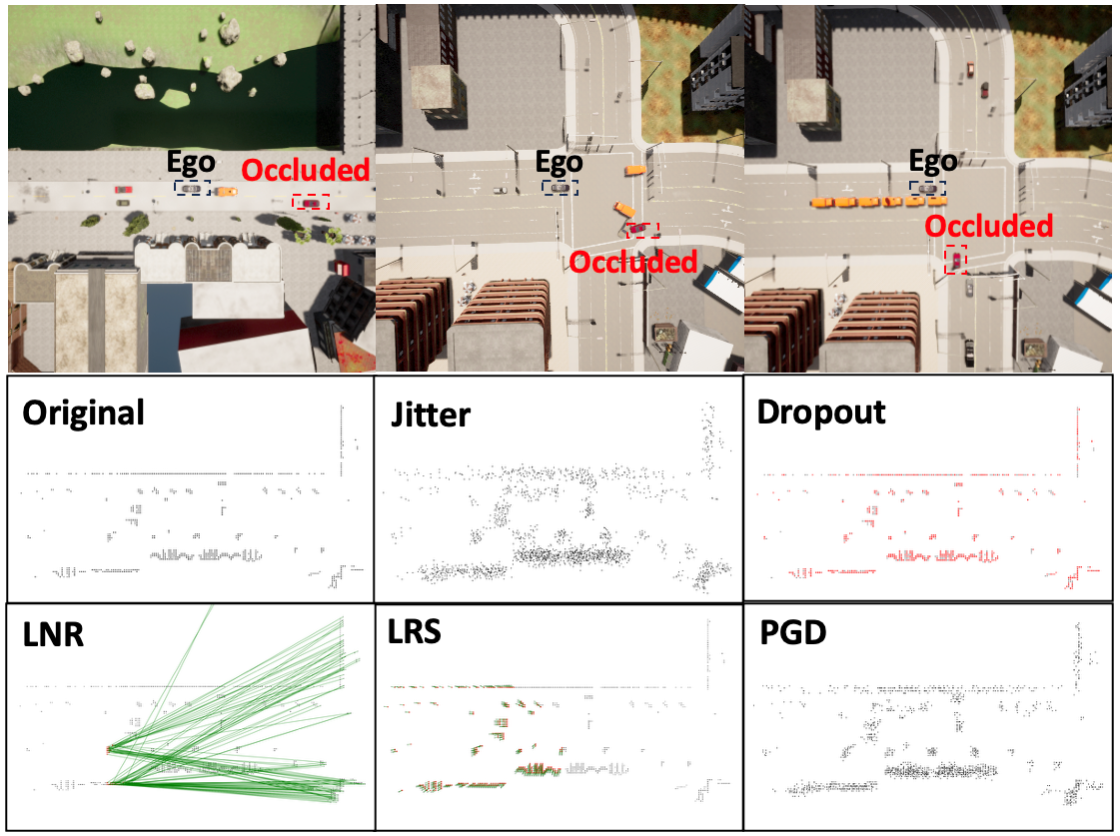}
    \caption{Illustration of scenarios and attacks in experiments. The top row shows BEVs of the three accident-prone scenarios with the highlight of ego vehicle and the occluded vehicle. The middle and bottom rows visualize the LiDAR point clouds under different conditions, including the original point cloud, point clouds corrupted by \textbf{Jitter} noise, \textbf{Dropout} (red dots indicate the original locations of removed points), \textbf{LNR} and \textbf{LRS} geometry manipulation attacks (green lines indicate the displacement of targeted points), and the gradient-based \textbf{PGD} attack. These attacks span different corruption patterns, increasing risks to safety-critical decision-making.}
    \label{fig:scenarios_attacks}
    \vspace{-5mm}
\end{figure}

We use the CARLA \cite{Dosovitskiy17} with the AutoCastSim \cite{autocast} to generate three connected autonomous driving (CAD) scenarios. The simulator contains one ego vehicle and $3-30$ collaborator vehicles. All the vehicles can share observations with each other once they reach a communication radius, which is set to $150$m. Each vehicle is mounted with a LiDAR sensor. The ground truth control of the ego vehicle can be directly obtained via the simulator. 
As shown in Figure ~\ref{fig:scenarios_attacks}, we set up three accident-prone scenarios including 
\begin{itemize}
    \item \textbf{Overtaking}: a truck occludes the ego vehicle's forward view on a two-lane road while oncoming traffic approaches from the opposite lane, requiring the ego vehicle to decide when to overtake safely.
    \item \textbf{Left Turn}: the ego vehicle attempts a left turn at an intersection where oncoming straight traffic is occluded by a vehicle ahead.
    \item \textbf{Red Light Violation}: the ego vehicle proceeds straight through an intersection while another vehicle runs the red light from a perpendicular direction, occluded by adjacent vehicles. 
\end{itemize}
All three scenarios feature occlusion-induced collision risks that demand timely braking decisions. 

In each scenario, we evaluate five attacks grouped into three categories, as shown in Figure ~\ref{fig:scenarios_attacks}, including 
\begin{itemize}
    \item \textbf{Stochastic Point Perturbations} globally affect all points, including 
    \textbf{Jitter \cite{ren2022benchmarking}}, which adds independent Gaussian noise ($\sigma = 1.0$\,m) to point coordinates, and 
    \textbf{Dropout \cite{ren2022benchmarking}}, which randomly removes 60\% of points and pads the cloud back to the original size by duplicating remaining points.
    
    \item \textbf{Geometry Manipulation Attacks} corrupt local structure in targeted regions, including 
    \textbf{LNR \cite{liu2024explicitly}} (Localized Neighbor Relocation), which identifies the densest azimuth sector ($60^\circ$) and relocates selected points and their neighbors to distant positions, and 
    \textbf{LRS \cite{liu2024explicitly}} (Localized Radial Shift), which shifts 65\% of points in the densest sector 1\,m radially outward.
    
    \item \textbf{Gradient-Based White-Box Attacks} include 
    \textbf{PGD}~\cite{madry2017towards}, which iteratively perturbs each point coordinate along the adversarial gradient to suppress the predicted braking probability, with a maximum perturbation of $\epsilon = 1.0$\,m per coordinate, a step size of $\alpha = 0.4$\,m per iteration, for a total of $3$ iterations.
\end{itemize}
All five attacks are applied to different targets, including \textbf{Ego} vehicle only, first \textbf{k} collaborators where $k=1,2,3$, and \textbf{All} vehicles including both ego and collaborators.

Each LiDAR point cloud is bounded to $x,y \in [-70,70]$m and $z \in [-2.5,-0.3]$m, voxelized at 0.5m resolution, and resampled to 2048 points. Local neighborhoods are constructed using $k{=}16$ nearest neighbors. We use PointTransformer\cite{zhao2021point} as the backbone, projecting 3D coordinates to a 32-dimensional feature space. Two downsampling stages reduce points by $4{\times}$ each while increasing feature dimensions to 128, producing 128 representative points per vehicle. Collaborator features are transformed into the ego frame, concatenated, voxel-pooled at 0.5,m, and resampled to match the ego points. The fused ego–collaborator points are further processed to obtain 64 final representative points, which are max-pooled and concatenated with an ego speed embedding before a prediction head outputs the brake logit.

We generate 12 trials for training and 12 trials for testing, where each trial corresponds to a continuous driving episode, yielding over 700 brake decisions for both training and testing. The training trials are randomly split into 80\% for training and 20\% for validation. 
The model is optimized using the Adam optimizer \cite{kingma2015adam} with a learning rate of $10^{-3}$ and weight decay of $10^{-5}$. Training is performed for up to 200 epochs with early stopping based on validation performance.

We further compare our RCDM method with one vanilla baseline and one existing method, including
\begin{itemize}\setlength{\itemsep}{0pt}\setlength{\parskip}{0pt}
    \item \textbf{Vanilla}: we create a baseline method that uses our RCDM without enabling reweighting module.
    \item \textbf{RoboSAC \cite{RoboSAC}}: a sampling-based consensus defense that uses the ego-only prediction as an anchor and accepts collaborator subsets within a consensus threshold. It requires that the ego observation is always reliable.
\end{itemize}
We follow the recent work \cite{liu2025mmcd}, which uses \textbf{ADR} (Accident Detection Rate) as the evaluation metric that is defined as the ratio of correct prediction over ground-truth. 
We also report corruption statistics to quantify attack intensity using two metrics. First, \textbf{chg} measures the percentage of points displaced beyond 0.25\,m. Second, \textbf{p95} denotes the 95th-percentile of point-wise displacement magnitude (in meters), capturing the typical upper bound of geometric distortion while being resilient to extreme outliers.

\begin{table}[b]
\centering
\caption{ \textbf{Summary of corruption statistics}. The combination of chg(\%) and p95 reveals distinct corruption patterns, ranging from overall displacement of points (Jitter, PGD) to local structured geometric distortion (LNR, LRS). Dropout will not introduce displacement of points.}
\label{tab:corruption_summary}
\footnotesize
\renewcommand{\arraystretch}{1.1}
\begin{tabular}{lcc}
\toprule
\textbf{Attack} & \textbf{chg (\%)} & \textbf{p95 (meter)} \\
\midrule
Jitter   & 93.7--95.8\% & 2.16--2.38 \\
Dropout  & --           & --         \\
LNR      & 5.1--8.6\%   & 0.31--0.74 \\
LRS      & 32.8--43.0\% & $\sim$1.00 \\
PGD      & 71.2--95.0\% & 1.17--1.35 \\
\bottomrule
\end{tabular}
\end{table}

\subsection{Experimental Results in CAD scenarios}
This experimental setting is particularly challenging due to the high intensity and diversity of corruption applied to ego-collaborator LiDAR observations. 
As shown in Table \ref{tab:corruption_summary}, under the Jitter and PGD attacks, more than 90\% of points are displaced beyond 0.25 m (e.g., chg\% up to 95.8\%), with 95th-percentile displacements exceeding 2.3 m, indicating severe global geometric distortion. Even localized attacks such as LNR introduce non-negligible perturbations, with p95 displacements reaching 0.7–0.74 m, which is sufficient to disrupt fine-grained geometric cues critical for braking decisions. Moreover, attacks targeting multiple collaborators (“All” or increasing $k$) compound these effects by simultaneously corrupting shared information, making it difficult to make decisions based solely on reliable observations.

Table~\ref{tab:full_adr_changed_p95} summarizes the quantitative performance of our RCDM compared with baseline and prior method based on ADR under five attacks and various target configurations. Given the quantitative results, we mainly answer these four questions.
\textbf{Q1} How effectively does RCDM perform under different attacks and target configurations?
Our RCDM consistently shows  the best performance on  ADR under all attack-target combinations across all three scenarios. Specifically,  PGD as the strongest attack causes the largest ADR drops in the vanilla model. Under the most severe setting (PGD applied to all vehicles), compared with the vanilla model,
our RCDM improves the performance on ADR from $0.68$ to $0.95$, $0.43$ to $0.68$ and $0.41$ to $0.47$ in three scenarios.
The consistent improvements indicate the effectiveness of our method under both individual and collaborative perceptual corruption.

\textbf{Q2} What are the benefits of RCDM compared with prior method?
Our RCDM consistently outperforms the prior work RoboSAC under all attack-target combinations across all three scenarios.
This is because RoboSAC always assumes a trustworthy ego vehicle to provide a reliable ego-centric anchor that enables effective incorporation of collaborators’ observations. However, this assumption no longer holds in our experimental setting. In addition, RoboSAC will reject all the collaborative information based on the thresholding of trust. In contrast, RCDM operates at the point level rather than rejecting the entire observation. It preserves useful points while downweighting only those points that deviate statistically from the group median.

\textbf{Q3} Why does RCDM show improvements even with original observations?
RCDM outperforms standard collaborative decision-making methods even in the absence of attacks because real-world LiDAR observations inherently contain inconsistencies due to occlusion and limited sampling density, which lead to inconsistent observations across robots. Conventional softmax-based attention assigns non-zero weights to all neighbors and therefore cannot fully suppress the influence of inconsistent collaborators. In contrast, RCDM adaptively down-weights these inconsistent neighbors through its reweighting mechanism, leading to more reliable information fusion and improved collaborative decision-making performance even under nominal, noise-only conditions.

\textbf{Q4} How resilient is RCDM to attacks targeting individual robots and collaborators?
While performance does not decrease monotonically as the attack scope expands from ego to collaborators and all robots, RCDM consistently outperforms the vanilla baseline across all attack targets.
In particular, RCDM maintains clear advantages under ego-only attacks and retains higher ADR than the vanilla baseline when collaborators or all robots are attacked. For example, under the PGD-All setting, RCDM achieves ADRs of 0.9481, 0.6765, and 0.4702 in three scenarios, whereas the vanilla baseline attains 0.6801, 0.4265, and 0.4107 in the same scenarios. This demonstrates improved tolerance to corrupted individual and collaborative observations.

\begin{table}[t]
\vspace{6pt}
\centering
\caption{\textbf{Quantitative results of RCDM under five attacks compared with prior and baseline methods based on ADR (accident detection rate) metrics in CAD scenarios}.
Bold indicates the best performance within each row.  {\em Ours consistently achieved the 
highest level of accuracy and robustness}, compared to other SOTA methods across all scenarios.}
\setlength{\tabcolsep}{2.5pt}
\renewcommand{\arraystretch}{1.08}
\resizebox{\columnwidth}{!}{%
\begin{tabular}{l l | ccc | ccc | ccc}
\toprule
\multirow{2}{*}{\textbf{Attack}} & \multirow{2}{*}{\textbf{Target}} &
\multicolumn{3}{c|}{\textbf{Overtaking}} &
\multicolumn{3}{c|}{\textbf{Left Turn}} &
\multicolumn{3}{c}{\textbf{Red Light Violation}}\\
\cmidrule(lr){3-5}\cmidrule(lr){6-8}\cmidrule(l){9-11}
& & Van. & RoboSAC & Ours & Van. & RoboSAC & Ours & Van. & RoboSAC & Ours \\
\midrule
Original & — &
0.9424 & — & \textbf{0.9798} &
0.7279 & — & \textbf{0.8309} &
0.5357 & — & \textbf{0.6012} \\
\midrule
\multirow{5}{*}{\textbf{Jitter}} & Ego &
0.9366 & — & \textbf{0.9741} &
0.5809 & — & \textbf{0.8162} &
0.5952 & — & \textbf{0.6905} \\
& k=1 &
0.9395 & 0.5706 & \textbf{0.9798} &
0.6912 & 0.7279 & \textbf{0.8088} &
0.5179 & 0.4643 & \textbf{0.5714} \\
& k=2 &
0.9395 & 0.5793 & \textbf{0.9798} &
0.6838 & 0.7059 & \textbf{0.8456} &
0.4821 & 0.4167 & \textbf{0.5476} \\
& k=3 &
0.9395 & 0.5850 & \textbf{0.9798} &
0.6912 & 0.7132 & \textbf{0.8015} &
0.4821 & 0.3988 & \textbf{0.5417} \\
& All &
0.9164 & — & \textbf{0.9683} &
0.6250 & — & \textbf{0.7941} &
0.5774 & — & \textbf{0.6190} \\
\midrule
\multirow{5}{*}{\textbf{Dropout}} & Ego &
0.9481 & — & \textbf{0.9712} &
0.6397 & — & \textbf{0.7574} &
0.5595 & — & \textbf{0.6369} \\
& k=1 &
0.9366 & 0.5648 & \textbf{0.9856} &
0.6838 & 0.7206 & \textbf{0.8015} &
0.5476 & 0.4405 & \textbf{0.5833} \\
& k=2 &
0.9135 & 0.5879 & \textbf{0.9769} &
0.6765 & 0.6912 & \textbf{0.8309} &
0.5536 & 0.4405 & \textbf{0.6131} \\
& k=3 &
0.9107 & 0.5793 & \textbf{0.9741} &
0.6838 & 0.6985 & \textbf{0.7868} &
0.5774 & 0.4464 & \textbf{0.6488} \\
& All &
0.9280 & — & \textbf{0.9654} &
0.6176 & — & \textbf{0.7574} &
0.5893 & — & \textbf{0.6667} \\
\midrule
\multirow{5}{*}{\textbf{LNR}} & Ego &
0.9107 & — & \textbf{0.9452} &
0.6618 & — & \textbf{0.7794} &
0.5000 & — & \textbf{0.5655} \\
& k=1 &
0.9424 & 0.5879 & \textbf{0.9769} &
0.6691 & 0.7279 & \textbf{0.8015} &
0.4940 & 0.4524 & \textbf{0.5655} \\
& k=2 &
0.9251 & 0.5850 & \textbf{0.9769} &
0.6985 & 0.7132 & \textbf{0.8162} &
0.4524 & 0.4226 & \textbf{0.5714} \\
& k=3 &
0.9193 & 0.5735 & \textbf{0.9798} &
0.6912 & 0.7132 & \textbf{0.8309} &
0.4464 & 0.4048 & \textbf{0.5595} \\
& All &
0.8818 & — & \textbf{0.9424} &
0.6838 & — & \textbf{0.8162} &
0.4167 & — & \textbf{0.5000} \\
\midrule
\multirow{5}{*}{\textbf{LRS}} & Ego &
0.9251 & — & \textbf{0.9798} &
0.6618 & — & \textbf{0.8162} &
0.4940 & — & \textbf{0.5655} \\
& k=1 &
0.9481 & 0.5850 & \textbf{0.9798} &
0.6985 & 0.6912 & \textbf{0.8235} &
0.5298 & 0.4524 & \textbf{0.5774} \\
& k=2 &
0.9395 & 0.5764 & \textbf{0.9798} &
0.7206 & 0.7279 & \textbf{0.8309} &
0.5179 & 0.4345 & \textbf{0.5357} \\
& k=3 &
0.9481 & 0.5850 & \textbf{0.9827} &
0.6985 & 0.7279 & \textbf{0.8162} &
0.4881 & 0.4583 & \textbf{0.5714} \\
& All &
0.9251 & — & \textbf{0.9712} &
0.6838 & — & \textbf{0.8235} &
0.5298 & — & \textbf{0.5774} \\
\midrule
\multirow{5}{*}{\textbf{PGD}} & Ego &
0.8617 & — & \textbf{0.9625} &
0.4632 & — & \textbf{0.6397} &
0.4583 & — & \textbf{0.5536} \\
& k=1 &
0.9020 & 0.6167 & \textbf{0.9741} &
0.6838 & 0.7279 & \textbf{0.7868} &
0.5119 & 0.4226 & \textbf{0.5833} \\
& k=2 &
0.8300 & 0.5908 & \textbf{0.9712} &
0.7206 & 0.7059 & \textbf{0.8235} &
0.4524 & 0.4226 & \textbf{0.4940} \\
& k=3 &
0.8184 & 0.5764 & \textbf{0.9654} &
0.6838 & 0.7279 & \textbf{0.8088} &
0.4464 & 0.4107 & \textbf{0.5476} \\
& All &
0.6801 & — & \textbf{0.9481} &
0.4265 & — & \textbf{0.6765} &
0.4107 & — & \textbf{0.4702} \\
\bottomrule
\end{tabular}}
\label{tab:full_adr_changed_p95}
\vspace{-4.3mm}
\end{table}

\begin{figure}[htbp]
    \centering
    \vspace{2mm}\includegraphics[width=\columnwidth]{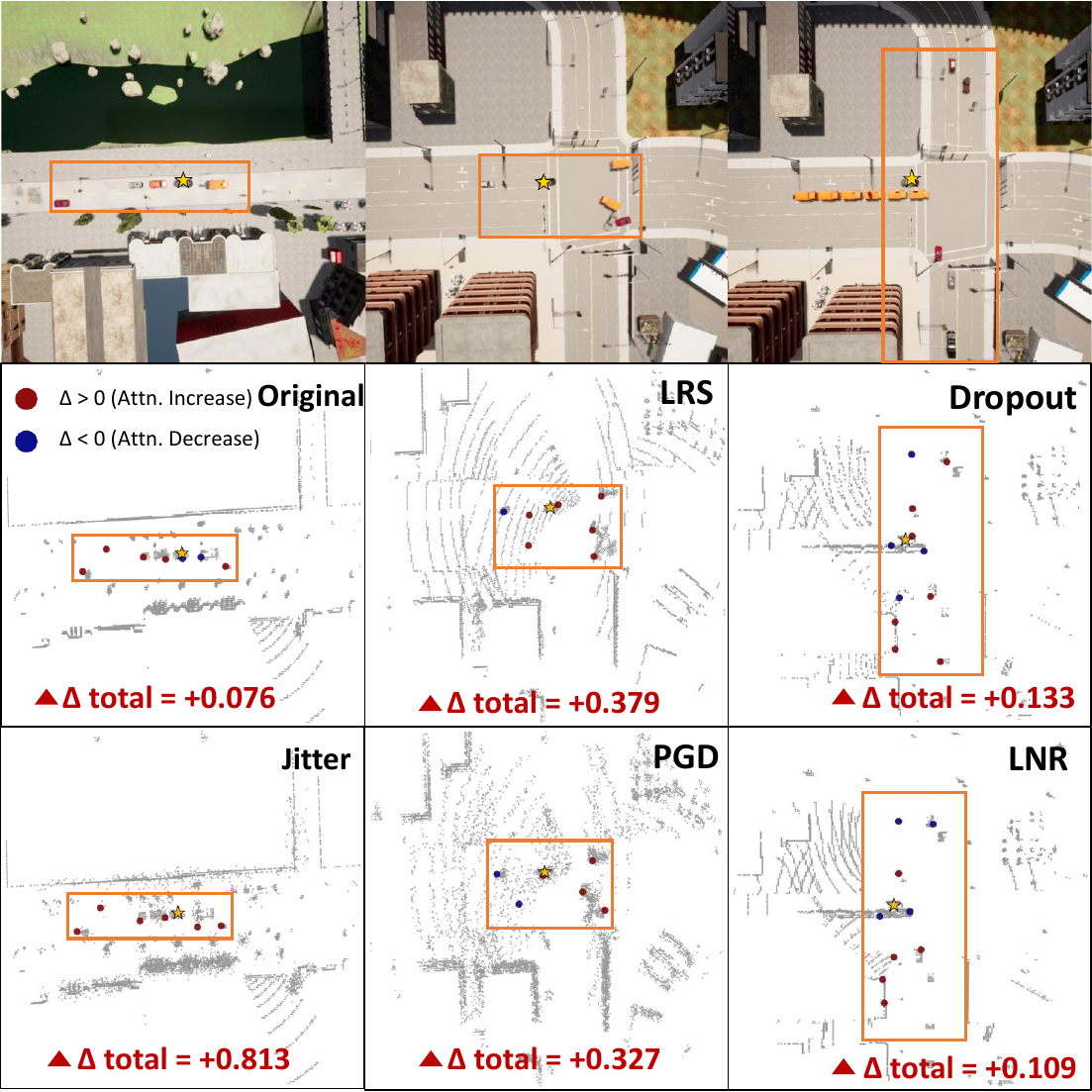}
    \caption{
    The visualization of attention changes of downsampled LiDAR points under various types of corruption.
    The first row shows the ROI in the red bounding box. 
    The second and third rows show the attention changes after applying RCDM.
    Star markers indicate the ego vehicle.
    Points in red and blue denote the increase and decrease of attention separately.
    Overall, RCDM increases attention in the safety-critical region under all corruptions.
}
    \label{fig:diss}
    \vspace{-6mm}
\end{figure}

\subsection{Discussion}

In Figure \ref{fig:diss}, we analyze how RCDM reweights attention under corrupted observations by visualizing the attention difference of downsampled points before and after applying our reweighting module. The attention difference is computed as
$\Delta = \mathrm{Attn}_{\text{RCDM}} - \mathrm{Attn}_{\text{vanilla}}$, where 
$\Delta$ quantifies how the introduction of reweighting module alters the relative importance assigned to each ego or collaborator observation during collaborative decision making. 

The visualization provides consistent evidence that this reweighting module is effective under diverse attacks. Across {Noise}, {LRS}, {Dropout}, {Jitter}, {PGD} and {LNR} attacks, RCDM consistently reduces attention on corrupted points (blue circles) while increasing attention on consistent points (red crosses) within the region of interest (ROI). This behavior is further quantified by the positive total attention difference $\Delta_{\text{total}}$ by adding up all the attention difference within the ROI, with values of $+0.076$, $+0.379$, $+0.133$, $+0.813$, $+0.327$, and $+0.109$. These results  demonstrate that RCDM effectively mitigates the influence of perceptual corruption.

\newcommand{\m}[3]{\makecell{#1\\ \scriptsize(#2,\ #3)}}



\section{Conclusion}
In this paper, we introduce RCDM that is a novel resilient collaborative decision-making method designed to handle corruption in both individual and collaborative perception. RCDM consists of an attention-based encoder for extracting
individual robot perceptual embeddings and an attention-
based decoder for fusing collaborator perceptions and making
decisions. To improve resilience to corrupted observations,
we design a novel plug-and-play reweighting module that
downweights the influence of corruption by analyzing
the consistency of neighborhood points relative to the local
structure and assigning smaller weights to points that deviate
strongly from the local median. Extensive experiments in high-fidelity simulations with diverse attacks and accident-prone scenarios demonstrate that RCDM achieves state-of-the-art resilient performance
under corrupted observations.

Our approach presents several limitations that suggest directions for future research. First, while RCDM has been validated in collaborative decision-making among connected vehicles, extending the framework to broader multi-robot settings, such as aerial–ground or heterogeneous robot teams, remains an open challenge. Second, the current design focuses on LiDAR-based perception. Integrating additional modalities such as camera, radar, or V2X signals could further enhance resilience and generalization in complex environments.

\newpage
\bibliographystyle{IEEEtran}
\bibliography{IEEEfull,reference}

\end{document}